# Enhancing Mortality Prediction in Heart Failure Patients: Exploring Preprocessing Methods for Imbalanced Clinical Datasets

Hanif Kia
Department of Biomedical Engineering,
K. N. Toosi University of Technology
Tehran, Iran
kia.hanif@email.kntu.ac.ir

Mansour Vali
Department of Biomedical Engineering,
K. N. Toosi University of Technology
Tehran, Iran
mansour.vali@eetd.kntu.ac.ir

Hadi Sabahi
Department of Biomedical Engineering,
K. N. Toosi University of Technology
Tehran, Iran
sabahihadi@email.kntu.ac.ir

*Abstract*— Heart failure (HF) is a critical condition in which the accurate prediction of mortality plays a vital role in guiding patient management decisions. However, clinical datasets used for mortality prediction in HF often suffer from an imbalanced distribution of classes, posing significant challenges. In this paper, we explore preprocessing methods for enhancing one-month mortality prediction in HF patients. We present a comprehensive preprocessing framework including scaling, outliers processing and resampling as key techniques. We also employed an aware encoding approach to effectively handle missing values in clinical datasets. Our study utilizes a comprehensive dataset from the Persian Registry Of cardio Vascular disease (PROVE) with a significant class imbalance. By leveraging appropriate preprocessing techniques and Machine Learning (ML) algorithms, we aim to improve mortality prediction performance for HF patients. The results reveal an average enhancement of approximately 3.6% in F1 score and 2.7% in MCC for tree-based models, specifically Random Forest (RF) and XGBoost (XGB). This demonstrates the efficiency of our preprocessing approach in effectively handling Imbalanced Clinical Datasets (ICD). Our findings hold promise in guiding healthcare professionals to make informed decisions and improve patient outcomes in HF management.

## I. Introduction

Heart Failure (HF) is a prevalent and serious medical condition that leads to significant hospitalization and mortality rates worldwide. Despite advancements in diagnosis and treatment, HF continues to impose a substantial burden on healthcare providers, making accurate mortality prediction of paramount importance [1]. Extensive evidence highlights that a considerable proportion of hospitalized patients with HF, approximately 40%, experience mortality or require readmission within a year, emphasizing the critical nature and ongoing challenges associated with this condition [2]. Early identification of risk factors is vital for improving prognosis and aiding in decision-making for patients with HF. Age and depression have been recognized as clinical predictors associated with increased mortality in HF [3].

Predicting mortality in HF patients is of utmost importance to identify high-risk individuals and optimize treatment strategies. However, accurate mortality prediction remains a complex task, particularly when working with Imbalanced Clinical Datasets (ICD). Imbalanced datasets are characterized by a significant disparity in the number of samples between the minority (e.g., deceased patients) and majority (e.g., surviving patients) classes.

In recent years, Machine Learning (ML) techniques have emerged as valuable tools for predicting in-hospital mortality among hospitalized patients. Chicco et al. applied different ML methods to predict survival in HF patients using serum creatinine and ejection fraction as the only features. The results showed that these two features alone were sufficient for accurate predictions, outperforming the original dataset features [4]. Eric D. Adler et al. employed a novel ML analytics to develop a risk score for predicting mortality in HF patients. The derived risk score, based on eight variables, demonstrated excellent discriminatory power in assessing mortality risk with an AUC of 0.88 [5]. Kwon et al. used a Deep Learning approach to predict in-hospital, 12-months and 36 months mortality and report AUC 0.88, 0.73 and 0.81 respectively [6]. Angraal et al, utilized Five different ML methods to develop models for predicting mortality and hospitalization in HF patients. The Random Forest (RF) model demonstrated the best performance, with a mean C-statistic of 0.72 for mortality prediction and 0.76 for hospitalization prediction [7].

Registry systems often encounter imbalanced classes, which can lead to low sensitivity when applying conventional ML algorithms. Various approaches address imbalanced classification problems, including oversampling and undersampling techniques that aim to improve classification performance [8]. Undersampling combined with ensemble methods is effective for imbalanced classification, but further improvements are needed for optimal performance in registry datasets.

Several studies are conducted in resent years which used resampling methods to enhance mortality prediction. A comparative analysis of ML classifiers was conducted by Ishaq et al. including tree-based methods. They addressed imbalance class problem by Synthetic Minority Oversampling Technique (SMOTE). The RF classifier achieved the highest F1-Score of 0.85 and performed well across all evaluation metrics [9]. In [10] Bokhare et al. proposed an MLP-SMOTE model using TensorFlow for predicting HF with an accuracy of approximately 91.55%. In [11] Aditya Sengupta et al. used a Logistic Regression (LR) to develop a weighted risk score for predicting post-discharge mortality at 1 year in patients undergoing heart operations and perform internal validation using a bootstrap-resampling approach. The risk score model demonstrated a C-statistic of 0.82 (95% confidence interval, 0.80-0.85).

In this study, we proposed a comprehensive framework to enhance mortality prediction in HF patients by exploring tailored preprocessing methods and model optimization techniques for ICD (Fig 1). Our research focused on key techniques, including Scaling, Resampling and Threshold Tuning to address class imbalance. By leveraging ML techniques such as RF, XGBoost Classifier (XGB), Easy Ensemble (EE), Support Vector Machine (SVM) and LR we aimed to advance risk stratification and enable personalized care for HF patients. Our study contributes valuable insights into preprocessing steps and ML approaches for accurate mortality prediction, fostering the development of CDSS and enhancing patient outcomes in the management of HF.

In the subsequent sections of this paper, we will provide an overview of the dataset and the features used in our study. We will then discuss the preprocessing techniques and ML methods employed to address the challenges in the dataset. Next, we will present the results obtained from our analysis and engage in a comprehensive discussion of their implications. Finally, the conclusion will be stated.

## II. DATASET AND FEATURES

The dataset used in this study consisted of records from patients hospitalized with decompensated or acute heart HF between March 2015 and October 2018. The data were extracted from the Persian Registry Of cardio Vascular diseasE (PROVE). This was initiated in Isfahan, Iran as the first registry program focused on cardiovascular diseases. The dataset encompassed the period before hospitalization, during hospitalization, and at the time of discharge, with follow-up data collected at 3, 6, and 12 months, as needed [12] [13].

A team of expert cardiologists selected 42 key features for current study, specifically focusing on before hospitalization and admission data, based on their clinical expertise and relevance. TABLE 1 provides a comprehensive overview of the dataset used in this study and selected features included in TABLE 2 were considered essential for predicting one-month mortality in HF patients.

## III. PREPROCESSING TECHNIQUES

Preprocessing is essential in analyzing clinical datasets, particularly for imbalanced classification tasks. This study utilized a comprehensive preprocessing pipeline to improve the reliability and performance of mortality prediction models in HF patients, preparing the dataset for analysis.

### A. Feature Scaling

Feature scaling is a preprocessing technique that aims to standardize the range and distribution of features in a dataset. It helps improve the performance of ML algorithms by allowing them to give appropriate importance to each feature, leading to more accurate predictions [14]. Three types of scalers are implemented in this study:

- Min-Max Scaler: It helps to maintain the relationships between data points, preserving the original distribution.

$$\hat{x}_i = \frac{x_i - x_{min}}{x_{max} - x_{min}} \qquad (1)$$

- Standard Scaler: It is highly beneficial when features have different scales, and some algorithms assume a Gaussian distribution for the input data.

$$\hat{x}_i = \frac{x_i - \mu}{\sigma} \qquad (2)$$

Where $\mu$ and $\sigma$ are mean and standard deviation of $x$ respectively.

- Robust Scaler: It is particularly useful for datasets with significant outliers, ensuring they do not overly impact the scaled values and offering a more dependable depiction of the data's distribution.

$$\hat{x}_i = \frac{x_i - Q_2(x)}{Q_3(x) - Q_1(x)} \qquad (3)$$

Where $Q_1(x)$, $Q_2(x)$ and $Q_3(x)$ are first, second and third quartile of $x$ respectively.

### B. Missing Data Treatment

In clinical datasets, missing values can occur due to various reasons, and their distribution can follow different patterns: Missing Completely at Random (MCAR), Missing at Random (MAR), and Not Missing at Random (NMAR) [15]. Each pattern requires specific handling methods to ensure accurate analysis and minimize potential biases [16] [17].

It is important to consider the missing data percentage and the relationship between missing and observed variables when determining if imputation is necessary. In cases where a high percentage of missing data exists or the validity of the available data is questionable, excluding the variable from the analysis may be a suitable approach [18].

Recent developments in handling missing values include imputation methods such as mean, median and mode or K-Nearest Neighbor (KNN) imputation and predict them. However, it is vital to consider the specific type of missingness, like MNAR and MAR, commonly found in clinical datasets. Traditional imputation methods may not be suitable for these cases, potentially leading to bias. Careful and appropriate handling of missing data is essential to maintain the reliability and accuracy of the analysis[19].

TABLE 1. Summary of Dataset and Patient Distribution

| Number of Records | 3892 |
|---|---|
| Number of Patients | 2888 |
| Age Distribution | Male: 68.97 (± 13.26) years<br>Female: 73.27 (± 11.66) years |
| Number of Features | 42 (6 Numerical, 36 Categorical) |
| Totall Number of Samples | 2477[a] |
| Class 0: Survived | 2141 |
| Class 1: Dead | 336 |

[a.] 411 patients were excluded due to the lack of information about the labels

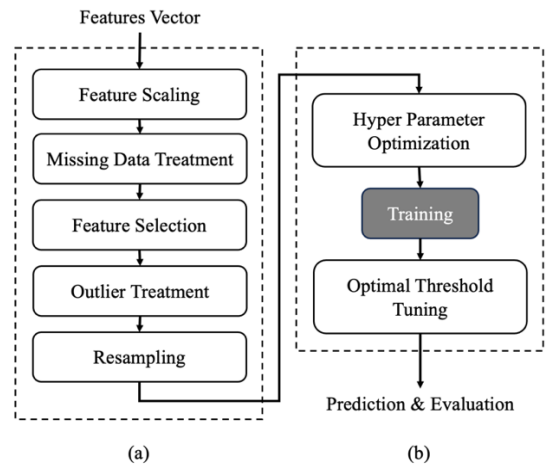

Fig 1. Comprehensive Framework for Handling Imbalanced Datasets. The framework consists of two main components: Preprocessing (a) and Model Optimization (b). All Processing steps are fitted on the training data and then applied to the validation data for evaluation.

In this study, a novel Missing Value Aware Encoding (MVAE) approach, which is depicted in Fig 2, was proposed to handle missing values by treating them as valuable information. It was observed that missing values, especially in categorical features, could indicate the absence of a specific attribute, thus representing meaningful patterns. To capture this information, a new entity was introduced specifically for missing values in categorical features, which could be encoded using either one-hot encoding or entity embedding techniques. This is a common method especially in the field of natural language processing [20]. For numerical features, missing values were represented as -1 after normalization between 0 and 1. By considering missing values as informative entities, the proposed method aimed to enable the model to extract patterns and make more accurate and reliable predictions.

*C. Feature Selection*

In this study, we utilize global and local feature selection methods to address the challenge of selecting informative features in clinical binary classification tasks. These methods, including Recursive Feature Elimination (RFE), Feature Importance based on Tree-based Models, and Laplacian Score, help us identify features with significant associations with the target variable while avoiding overfitting. RFE recursively removes less important features based on their importance rankings, ensuring the selection of the most informative ones for modeling [21]. Tree-based models like RF and Gradient Boosting provide feature importance scores, aiding in ranking features based on their discriminative power [22]. Additionally, the Laplacian Score methodology evaluates the relevance of features by considering the local structure of the data and its relationship to neighboring data points, contributing to improved classification performance [23].

*D. Outlier Detection and Treatment*

Outlier detection and treatment are important for learning from ICD. In this study, Local Outlier Factor (LOF) was used to detect outliers based on data point density.

LOF is an unsupervised outlier detection algorithm that assesses the local density of data points. It calculates a score for each data point based on its relative density compared to its neighboring points. Data points with significantly lower density scores are identified as potential outliers, providing insights into the local anomaly behavior of the dataset [24].

*E. Resampling Techniques*

Resampling techniques are pivotal in addressing the imbalanced nature of the labels in the clinical classification task [25]. In this study, SMOTETomek was employed as a resampling technique to balance the training data. This process involved two main steps: removing Tomek links and performing oversampling using SMOTE.

Tomek links are pairs of instances in close proximity from different classes, and removing them enhances separation between the minority and majority classes, improving discrimination and classification performance. SMOTE oversampling generates synthetic samples by interpolating neighboring feature vectors of the minority class, effectively increasing the number of minority samples and achieving a more balanced distribution in the training data. These steps address class imbalance and contribute to improved classification in the study [26] [27].

IV. MACHINE LEARNING TECHNIQUES

This section provides an overview of the ML techniques employed in our study. We discuss the distinct capabilities and methodologies of each algorithm and investigate their effectiveness in handling the imbalanced labels prevalent in clinical datasets.

*A. Classification Models*

RF is an ensemble learning algorithm that combines multiple decision trees for making predictions. It avoids overfitting by training each tree on a random subset of features and bootstrapped samples. RF could effectively handle imbalanced and non-scaled datasets [28].

XGB is a highly effective gradient boosting algorithm that combines weak models to correct errors sequentially. It is efficient, scalable, and incorporates regularization techniques for improved generalization [29]. In the context of imbalanced labels, XGB handles class imbalance by assigning a higher weight to the minority class, enhancing classification performance for the minority class [30].

EE is a powerful ensemble learning algorithm designed to address class imbalance in classification tasks. It starts by randomly under-sampling the majority class to create balanced training sets. Multiple classifiers, typically decision trees, are then trained on these subsets. During the prediction phase, each classifier's output is combined through voting or averaging to produce the final prediction [31].

SVM is robust ML algorithm which aims to find an optimal hyperplane that separates data points into different classes while maximizing the margin between the classes [32].

LR is a statistical modeling technique used for binary classification, which models the relationship between input variables and class probabilities using the logistic function. It offers advantages such as interpretable results, computational efficiency, and the ability to handle both linear and non-linear relationships between features and class probabilities.

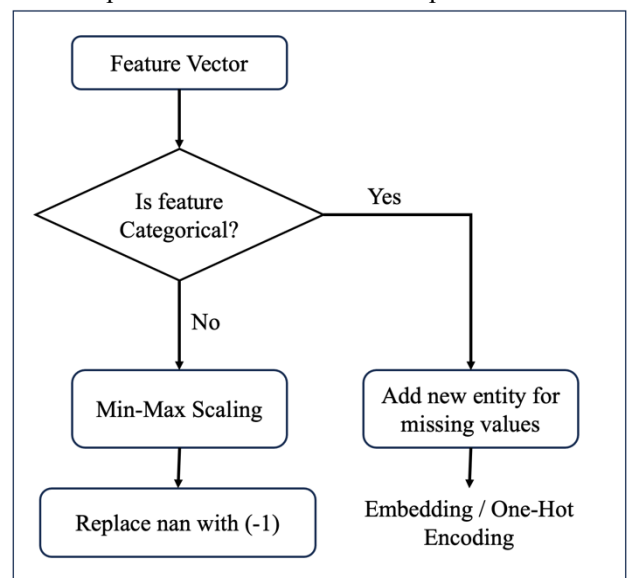

Fig 2. Missing Value Aware Encoding (MVAE). It replaces missing values of numerical features with -1 after normalizing between 0 and 1 and for categorical features with missing values, new entities are introduced to represent the missing entries.

## B. Hyperparameter Optimization

Hyperparameter optimization is a critical step in fine-tuning machine learning algorithms to achieve optimal performance. In the context of imbalanced datasets, this process becomes particularly important to ensure accurate classification of the minority class. Grid search, employed in this study, systematically explores the hyperparameter space, allowing us to identify the best combination that leads to improved performance, especially for imbalanced classes [33].

## C. Threshold Optimization

Threshold optimization is crucial for imbalanced classification tasks as it allows us to strike a balance between precision and recall. ROC curve analysis, which plots sensitivity against 1-specificity at various threshold values, is a powerful tool to identify the threshold that achieves the desired true positive rate, ensuring effective model performance [34]. In this study, final results were reported without threshold tuning (threshold=0.5).

TABLE 2. Features of the preprocessed HF registry data, presented as categorical variables (n (%)) and numerical variables (mean ± SD), organized into eight different groups. Significant statistical differences (p < 0.01) are denoted. Key features include Myocardial Infarction (MI), COPD (Chronic Obstructive Pulmonary Disease), SBP (Systolic Blood Pressure), DBP (Diastolic Blood Pressure), JVP (Jugular Venous Pulse), CPO (Cold Peripheral Organs), CABG (Coronary Artery Bypass Grafting), PCI (Percutaneous Coronary Intervention), CRT-D (Cardiac Resynchronization Therapy-Defibrillator), ICD (Implantable Cardioverter Defibrillator), NIV (Non-Invasive Ventilation), Hb (Hemoglobin), and BUN (Blood Urea Nitrogen).

| # | Group Features | Types | Features | | Class "0" n=2141 | Class "1" n=336 | P-value |
|---|---|---|---|---|---|---|---|
| 1 | Demographic | Cat | Index | | 399(18.6%) | 57(17.0%) | 0.5096 |
| 2 | Aetiologia | Cat | Primary Hypertension | | 1417(66.8%) | 208(62.7%) | 0.1569 |
| 3 | | Cat | Primary Heart Ischemic | | 1750(83.8%) | 257(81.6%) | 0.3624 |
| 4 | | Cat | Primary Valvular Heart | | 809(38.1%) | 163(48.9%) | 0.0002 |
| 5 | Medical History | Cat | Hypertension | | 1417(66.6%) | 209(62.8%) | 0.1906 |
| 6 | | Cat | Arrhythmia | | 409(19.1%) | 81(24.1%) | 0.0408 |
| 7 | | Cat | Diabetes | | 974(45.6%) | 159(47.5%) | 0.5736 |
| 8 | | Cat | COPD | | 286(13.4%) | 47(14.0%) | 0.8206 |
| 9 | | Cat | Thyroid | | 145(6.8%) | 35(10.4) | 0.023 |
| 10 | | Cat | Stroke | | 98(4.6%) | 25(7.4%) | 0.0353 |
| 11 | | Cat | Anemia | | 213(10.0%) | 86(25.7) | <0.001 |
| 12 | | Cat | Kidney Disease | | 501(23.6%) | 131(39.1%) | <0.001 |
| 13 | Vital Sign | Num | SBP | | 133.12(±28.91) | 115.19(±27.9) | <0.001 |
| 14 | | Num | DBP | | 81.81(±16.67) | 74.19(±14.64) | <0.001 |
| 15 | | Num | Heart Rate | | 88.06(±20.13) | 93.39(±27.2) | <0.001 |
| 16 | Physical Examination | Cat | Edema | | 979(62.2%) | 200(74.9%) | <0.001 |
| 17 | | Cat | Jvp | | 366(28.8%) | 87(40.3%) | <0.001 |
| 18 | | Cat | Crackle | | 1346(71.0%) | 254(81.4%) | <0.001 |
| 19 | | Cat | CPO | | 34(5.2%) | 28(28.6%) | <0.001 |
| 20 | Procedures | Cat | CABG | | 292(40.8%) | 44(41.1%) | 1 |
| 21 | | Cat | PCI | | 525(56.8%) | 51(42.9%) | 0.0054 |
| 22 | | Cat | CRT-D | | 28(1.3%) | 5(1.5%) | 0.9904 |
| 23 | | Cat | ICD | | 104(4.9%) | 18(5.4%) | 0.7965 |
| 24 | | Cat | Hemodialysis | | 49(2.3%) | 31(9.2%) | <0.001 |
| 25 | | Cat | NIV | | 44(2.1%) | 94(28.0&) | <0.001 |
| 26 | Medications | Cat | Captopril | | 276(13.6%) | 36(11.7%) | 0.4089 |
| 27 | | Cat | Losartan | | 905(44.6%) | 120(39.1%) | 0.0772 |
| 28 | | Cat | Metoral | | 973(48.0%) | 116(37.8%) | 0.001 |
| 29 | | Cat | Hydrochlorothiazide | | 112(5.5%) | 25(8.2) | 0.0889 |
| 30 | | Cat | Furosemide | | 885(43.7%) | 171(55.9%) | <0.001 |
| 31 | | Cat | Spironolactone | | 472(23.3%) | 88(28.8%) | 0.0457 |
| 32 | | Cat | Digitalis | | 498(24.6%) | 92(30.1%) | 0.0459 |
| 33 | | Cat | Atorvastatin | | 924(45.6%) | 107(34.9%) | <0.001 |
| 34 | | Cat | Nitrocountine | | 908(44.7%) | 116(37.9%) | 0.0293 |
| 35 | | Cat | Warfarin | | 379(18.7%) | 74(24.1%) | 0.0312 |
| 36 | | Cat | ASA | | 309(15.2%) | 42(13.6%) | 0.5393 |
| 37 | | Cat | Plavix | | 353(16.5%) | 36(10.7%) | 0.8624 |
| 38 | Biomarker | Num | Hb | | 13.09(±2.25) | 12.65(±2.53) | 0.0024 |
| 39 | | Num | Creatine | | 1.53(±1.46) | 2.09(±1.27) | <0.001 |
| 40 | | Num | Bun | | 27.28(±17.56) | 44.68(±26.74) | <0.001 |
| 41 | | Cat | Troponin | 0=Not Down | 358(25.1%) | 76(22.6%) | <0.001 |
| | | | | 1=Positive | 231(10.8%) | 98(29.1%) | |
| | | | | 2=Negative | 1550(72.4%) | 161(47.9%) | |
| 42 | Demographic | Cat | Sex | Male | 1297(60.9%) | 195(58.0%) | 0.3548 |
| | | | | Female | 834(38.0%) | 141(35.8%) | |

## V. EXPERIMENTAL DESIGN

The experimental design in this study (Fig 3) follows a systematic and sequential approach, utilizing 10-fold cross-validation with 3 repetitions for preprocessing the data. The step-by-step procedure includes scaling the data, handling missing values, outlier detection, and resampling, allowing for a thorough evaluation of their individual and cumulative effects on model performance. Models are further optimized using grid search and evaluated on a validation set, with resulting metrics stored in JSON format for comparison.

In evaluating the models, a comprehensive set of performance metrics is utilized, including:

$$Accuracy = \frac{(TP + TN)}{(TP + FP + TN + FN)} \quad (4)$$

$$F1\ Score = \frac{TP}{TP + \frac{1}{2}(FP + FN)} \quad (5)$$

$$Precision = \frac{TP}{TP + FP} \quad (6)$$

$$Recall = \frac{TP}{TP + FN} \quad (7)$$

Roc-Auc Score: Illustrating the model's ability to discriminate between the two classes.

Matthews Correlation Coefficient (MCC): Is a metric that measures the quality of binary classifications, considering true positive, true negative, false positive, and false negative rates, providing a balanced evaluation in imbalanced datasets.

$$MCC = \frac{TP \times TN - FP \times FN}{\sqrt{(TP + FP)(TP + FN)(TN + FP)(TN + FN)}} \quad (8)$$

Visualizations include two bar plots, focusing on F1 score and MCC, due to their balanced evaluation, to highlight the impact of preprocessing methods on the performance of the models. These plots provide a comprehensive overview of the classifiers' effectiveness across different sets of preprocessing methods, with detailed results gathered in a table for further comparison.

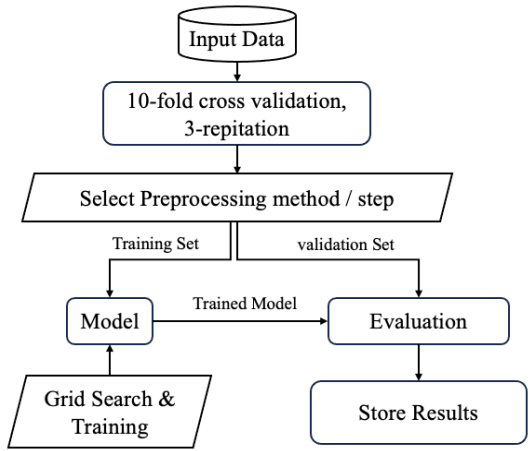

Fig 3. Experimental Workflow and Preprocessing Steps for Mortality Prediction in HF Patients.

## VI. RESULTS AND DISCUSSION

In this section, we present the results and discussion of our study, focusing on the performance evaluation of different methods for mortality prediction in HF patients. We analyze each method individually, examining their effectiveness in handling class imbalance and improving classification performance.

Fig 4 illustrates the impact of different scaling methods on classifier performance when applied exclusively to numerical features. The study transformed categorical features into one-hot encoding. It shows tree-based models are less sensitive to data scaling, while SVM model exhibit notable performance variations depending on the scaling method. Standard scaling outperforms other methods in these models, possibly due to the specific characteristics of their learning algorithms. However, the robust scaler does not yield satisfactory results in our completely imbalanced dataset with a significant number of outliers.

Various imputation methods for handling missing values were compared, including mode and mean filling for categorical and numerical features, respectively, along with KNN imputation for different k values. The proposed MVAE method, allowing the model to learn how to handle missing data, was also evaluated. Fig 5 demonstrates that the choice of imputation method has minimal impact on the average performance of different classifiers, with MVAE performing well in XGB and EE. However, SVM and LR do not perform well with MVAE. Given the overall similarity in results across methods in tree-based models, we recommend utilizing MVAE as it holds the potential to generate more reliable outcomes by enabling the model to effectively learn from missing data.

To ensure the continuity of our work, we standardized the preprocessing approach by utilizing the Standard Scaler and MVAE method as the primary techniques. These methods served as the foundation for obtaining the subsequent results, and their influence played a pivotal role in the outcomes. By fixing the scaler and missing value treatment, we established a consistent and reliable framework for evaluating the effectiveness of other preprocessing methods and their impact on the classification performance of various models.

We implemented various feature selection methods, including Laplacian Score and RF Importance, to select the top 70% informative features. Additionally, the Recursive Feature Elimination (RFE) method with F1 score as the criterion was employed for feature selection across different classifiers. Feature selection was conducted before converting categorical features to one-hot encoding. Fig 6 illustrates that RFE had a notable impact on the F1 score of RF signifying its effectiveness in classification (P-value=0.029). Although our study initially involved 42 expert-selected features, feature selection proved crucial when working with all features. In the final step, we utilized the RFE method through multiple runs using different estimators we identified 31 features out of the initial set of 42. Feature selection plays a vital role in preprocessing clinical datasets, especially in imbalanced datasets, as supported by extensive research.

Outliers can have a substantial impact on the training process of many models in the training dataset. While various approaches, such as removing or imputing outliers, are commonly used, dealing with outliers in ICD requires careful consideration. Since these datasets often represent individuals with specific and severe diseases, the presence of outliers is expected, and their removal may not be advisable. Models with robust algorithms to outliers, such as RF and XGB, can be employed to handle outliers effectively.

Fig 7 illustrates the impact of removing outliers detected by the LOF with different number of neighbors on classifier performance. The results indicate that removing outliers in tree-based models does not lead to significant changes in terms of F1-score. However, for XGB, the MCC bars show that the overall classifier performance can be enhanced (P_value = 0.031) by running the algorithm with 2 neighbors (n=2). Conversely, for parametric models, the changes that lead to enhancements are not statistically significant, and in the case of the SVM model, the removal of samples has resulted in a reduction in classification quality. As previously mentioned, these results are influenced by the Standard Scaler, and it is evident that the detection and removal of outliers after standardization have a negligible effect on classification quality in our case. Considering the specific nature of ICD and the potential benefits outliers provide in tree-based models, cautious consideration should be given when deciding whether to remove outliers or retain them in the training process.

Resampling techniques have proven effective in addressing imbalanced datasets by enhancing the detection of true positives and improving the performance of classifiers. In this study, attaining a resampling ratio between 0.5 to 0.7 (mu) led to improved outcomes without significantly altering the discrimination capability (The ratio of the minority class to the majority class before resampling was approximately 0.16).

$$mu = \frac{n\_minority_{after\ resampling}}{n\_majority_{after\ resampling}} \quad (9)$$

Fig 8 shows that resampling leads to an increase in the F1-score across almost all models, indicating enhanced true positive detection capabilities. Furthermore, the MCC in the RF and EE models shows a significant improvement, reflecting an overall enhancement in classification quality. The optimal mu value may vary for different models and datasets, necessitating a trial-and-error approach in selecting the most suitable resampling method and mu value. For RF and EE, mu=0.7 proves to be appropriate. Notably, the XGB model, with its ability to assign weights to the minority class, performs robustly on imbalanced datasets, as evidenced by its relatively minimal response to resampling in this study.

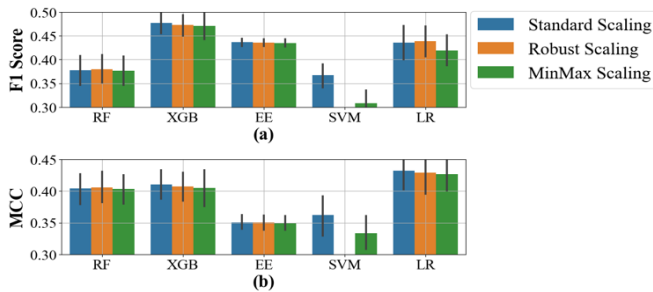

Fig 4. Summarizes the effect of different scaling methods on classifier performance in terms of F1-Score (a) and MCC (b). The results were generated after filling missing values with the mode for categorical features and the mean for numerical features.

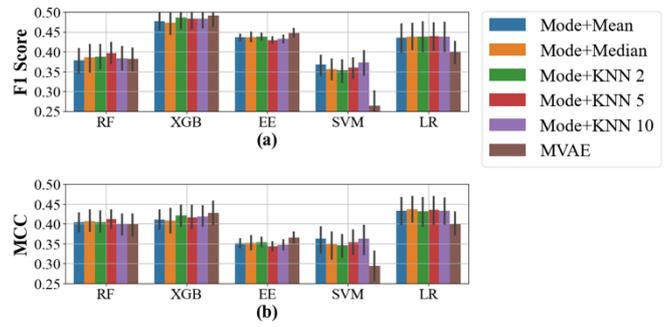

Fig 5. Summarizes the effect of different imputation methods on classifier performance in terms of F1-Score (a) and MCC (b).

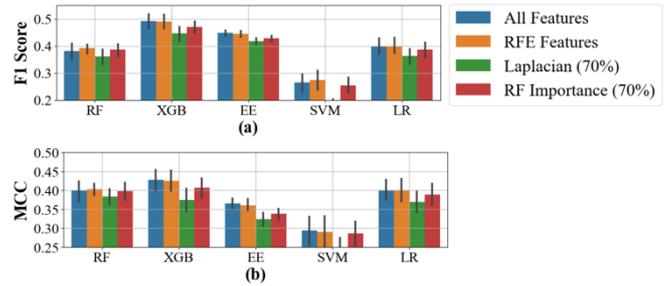

Fig 6. Summarizes the effect of different feature selection methods on classifier performance in terms of F1-Score (a) and MCC (b).

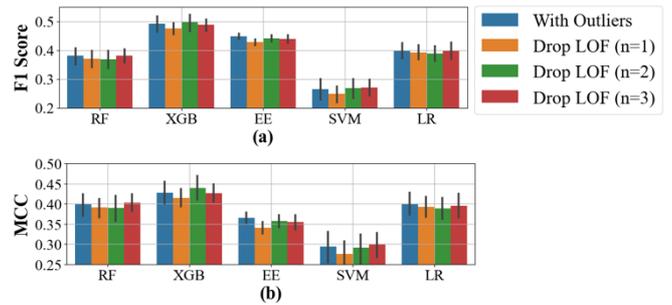

Fig 7. Summarizes the impact of outlier detection and removal using the LOF method with varying neighborhood numbers (n) on classifier performance in terms of F1-Score (a) and MCC (b).

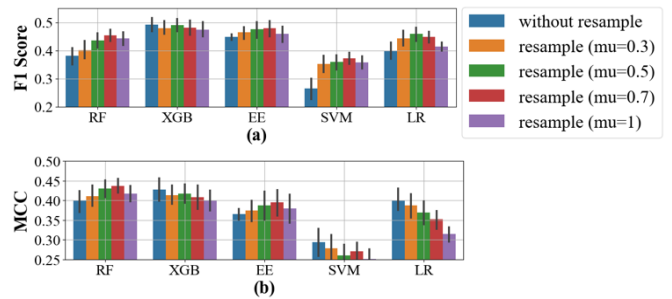

Fig 8. Summarizes the impact of SMOTETomek resampling with varying sampling ratio (mu) on classifier performance in terms of F1-Score (a) and MCC (b).

In this study, we systematically explored the impact of four preprocessing sets on the classification performance for one-month mortality prediction in HF patients. The preprocessing sets were defined as follows:

- SET 1: MVAE + Standard Scaling
- SET 2: SET 1 + RFE + Outlier Removal (LOF with 2 neighbors)
- SET 3: SET 2 + SMOTETomek (mu=0.7)
- SET 4: SET 1 + RFE + SMOTETomek (mu=0.7), Note: Outliers are not dropped in SET 4, affecting both resampling and final model performances.

To emphasize the effectiveness of the MVAE method, which inherently includes a Min-Max Scaler, we applied the MVAE before standard scaling in our preprocessing pipeline.

The results presented in TABLE 3 provide a comprehensive comparison of the classification performance across the three preprocessing sets. Additionally, Fig 9 presents box plots depicting the distribution of different metrics, providing further insights into the classifiers' performance.

Notably, we observed significant enhancements of 3.6% in F1-score and 2.7% in MCC for the average results of tree-based models, which include RF, XGB, and EE. These improvements were achieved by comparing the performance of Set 1 (baseline) with that of Set 4, which was proposed in this study. Set 4 involved a combination of MVAE with Standard Scaling, RFE as feature selection, and SMOTETomek resampling (mu=0.7). The notable improvements in F1-score and MCC demonstrate the effectiveness of our proposed preprocessing approach in enhancing the performance of tree-based classifiers for predicting one-month mortality in HF patients. These findings have significant implications for improving medical diagnosis and decision-making processes.

## VII. FUTURE DIRECTIONS

Future studies in this field may focus on refining feature engineering and selection methods to improve classification performance. Additionally, exploring ensemble and hybrid models specifically designed for ICD could lead to more robust and accurate results. Exploring hybrid and combined resampling techniques may also lead to more effective strategies for handling class imbalance. Moreover, incorporating deep learning methods, such as Convolutional Neural Networks (CNN) and Recurrent Neural Networks (RNN), for sequence analysis of patients' medical history could provide valuable insights for disease management and CDSS.

## VIII. CONCLUSION

In conclusion, this study focused on predicting one-month mortality in patients with HF using ML techniques. Through extensive experimentation and analysis, we evaluated various classifiers, scaling methods, outlier handling approaches, missing values treatment, and resampling techniques. Our findings emphasize the effectiveness of tree-based models in handling ICD and significantly improving predictive accuracy. Specifically, the proposed preprocessing steps demonstrated the potential to enhance the F1-score and MCC of tree-based models by up to 4% and 3%, respectively. Suitable ML algorithms and preprocessing strategies can enhance the accuracy of CDSS and disease management processes, providing a foundation for future research and improving patient outcomes in healthcare practices.

## IX. ACKNOWLEDGEMENT

We extend our heartfelt gratitude to the Cardiovascular Research Institute of Isfahan University of Medical Science for their invaluable support and data, which made this paper possible. Their commitment to advancing cardiovascular research is deeply appreciated.

TABLE 3. Performance Comparison of Different Classifiers for One-Month Mortality Prediction in HF Patients Across Different Preprocessing Sets. Set1: MVAE + Standard Scaling. Set2: Set1 + RFE + Outlier Removal (LOF with 2 neighbors). Set3: Set2 + SMOTETomek (mu=0.7). Set4: Set1 + RFE + SMOTETomek (mu=0.7) - Note: Outliers are not dropped in Set4, affecting both resampling and final performances.

| Preprocessing Methods | model | Accuracy | F1 Score | Precision | Recall | Roc-Auc | MCC |
|---|---|---|---|---|---|---|---|
| SET 1 | RF | **88.8 (±0.6)** | 38.2 (±5.6) | **76.3 (±7.1)** | 25.7 (±4.6) | **83.0 (±1.8)** | 39.9 (±5.3) |
| | XGB | 87.7 (±1.0) | **49.2 (±5.0)** | 55.8 (±4.9) | 44.1 (±5.4) | 81.4 (±1.9) | **42.7 (±5.4)** |
| | EE | 75.3 (±1.6) | 44.8 (±2.0) | 32.2 (±1.7) | **73.7 (±4.6)** | 80.0 (±2.4) | 36.5 (±2.7) |
| | SVM | 87.6 (±0.8) | 26.5 (±7.2) | 69.1 (±13.4) | 16.7 (±5.1) | 76.9 (±2.6) | 29.4 (±7.3) |
| | LR | 88.7 (±0.8) | 39.9 (±5.4) | 72.0 (±7.3) | 27.8 (±4.7) | 78.7 (±2.6) | 40.0 (±5.5) |
| SET 2 | RF | **88.7 (±0.8)** | 36.9 (±6.0) | **76.6 (±8.9)** | 24.5 (±4.6) | **82.8 (±2.0)** | 39.0 (±6.1) |
| | XGB | 88.2 (±1.1) | **48.7 (±5.3)** | 59.0 (±5.7) | 43.2 (±5.6) | 80.6 (±2.1) | **43.0 (±5.6)** |
| | EE | 74.7 (±1.5) | 44.1 (±2.1) | 31.6 (±1.6) | **73.5 (±5.1)** | 80.3 (±2.5) | 35.7 (±3.0) |
| | SVM | 87.6 (±0.8) | 26.9 (±6.4) | 66.6 (±12.6) | 17.0 (±4.4) | 76.8 (±2.5) | 29.2 (±7.4) |
| | LR | 88.5 (±0.8) | 38.9 (±5.0) | 70.6 (±8.0) | 27.0 (±4.2) | 78.6 (±2.6) | 38.8 (±5.4) |
| SET 3 | RF | **89.0 (±0.6)** | 42.1 (±5.0) | **73.2 (±5.8)** | 29.8 (±4.5) | **82.8 (±1.7)** | 42.0 (±4.6) |
| | XGB | 87.3 (±1.3) | **49.3 (±4.7)** | 54.0 (±6.1) | 45.6 (±5.2) | 80.7 (±1.5) | **42.4 (±5.3)** |
| | EE | 86.5 (±1.5) | 48.6 (±5.8) | 50.4 (±6.2) | **47.1 (±6.2)** | 81.6 (±2.3) | 41.0 (±6.6) |
| | SVM | 84.0 (±1.7) | 41.0 (±5.6) | 41.3 (±5.6) | 41.1 (±6.8) | 75.6 (±2.3) | 31.9 (±6.3) |
| | LR | 81.6 (±1.8) | 45.4 (±3.7) | 38.1 (±3.8) | 56.2 (±4.3) | 78.9 (±2.0) | 35.8 (±4.6) |
| SET 4 | RF | **89.0 (±0.6)** | 45.4 (±4.0) | **72.7 (±6.4)** | 31.7 (±4.3) | **82.5 (±1.9)** | 43.7 (±3.6) |
| | XGB | 88.8 (±1.2) | **50.1 (±5.1)** | 57.7 (±6.4) | 45.4 (±6.3) | 81.5 (±2.2) | **44.1 (±5.7)** |
| | EE | 85.5 (±1.4) | 47.6 (±5.7) | 42.3 (±5.3) | 54.8 (±7.7) | 81.4 (±2.0) | 39.5 (±6.1) |
| | SVM | 82.2 (±1.3) | 37.3 (±3.8) | 35.9 (±3.8) | 39.0 (±4.9) | 72.0 (±3.1) | 27.1 (±4.4) |
| | LR | 81.2 (±2.1) | 44.9 (±3.7) | 37.5 (±4.0) | **56.1 (±4.1)** | 78.5 (±2.0) | 35.2 (±4.5) |

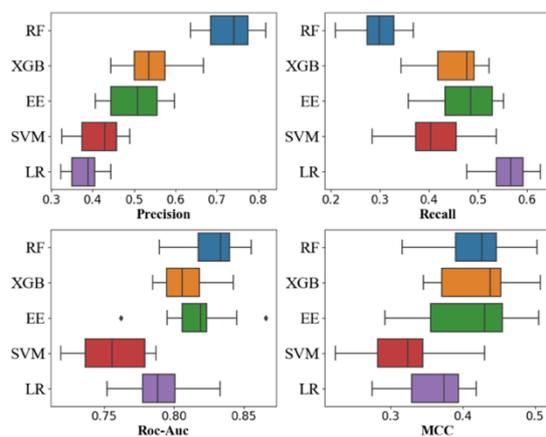

Fig 9. Box plots comparing different performance metrics for various classifiers in Set4.

## REFERENCES

[1] Yancy, Clyde W., et al. "2013 ACCF/AHA guideline for the management of heart failure: a report of the American College of Cardiology Foundation/American Heart Association Task Force on Practice Guidelines." Journal of the American college of cardiology 62.16 (2013): e147-e239.

[2] Pocock SJ, Wang D, Pfeffer MA, Yusuf S, McMurray JJ, Swedberg KB, Ostergren J, Michelson EL, Pieper KS, Granger CB. Predictors of mortality and morbidity in patients with chronic heart failure. European heart journal. 2006 Jan 1;27(1):65-75.

[3] Mahmood SS, Levy D, Vasan RS, Wang TJ. The Framingham Heart Study and the epidemiology of cardiovascular disease: a historical perspective. The lancet. 2014 Mar 15;383(9921):999-1008.

[4] Chicco D, Jurman G. Machine learning can predict survival of patients with heart failure from serum creatinine and ejection fraction alone. BMC medical informatics and decision making. 2020 Dec;20(1):1-6.

[5] Adler ED, Voors AA, Klein L, Macheret F, Braun OO, Urey MA, Zhu W, Sama I, Tadel M, Campagnari C, Greenberg B. Improving risk prediction in heart failure using machine learning. European journal of heart failure. 2020 Jan;22(1):139-47.

[6] Kwon JM, Kim KH, Jeon KH, Lee SE, Lee HY, Cho HJ, Choi JO, Jeon ES, Kim MS, Kim JJ, Hwang KK. Artificial intelligence algorithm for predicting mortality of patients with acute heart failure. PloS one. 2019 Jul 8;14(7):e0219302.

[7] Angraal S, Mortazavi BJ, Gupta A, Khera R, Ahmad T, Desai NR, Jacoby DL, Masoudi FA, Spertus JA, Krumholz HM. Machine learning prediction of mortality and hospitalization in heart failure with preserved ejection fraction. JACC: Heart Failure. 2020 Jan;8(1):12-21.

[8] Wallace BC, Small K, Brodley CE, Trikalinos TA. Class imbalance, redux. In2011 IEEE 11th international conference on data mining 2011 Dec 11 (pp. 754-763). Ieee.

[9] Ishaq A, Sadiq S, Umer M, Ullah S, Mirjalili S, Rupapara V, Nappi M. Improving the prediction of heart failure patients' survival using SMOTE and effective data mining techniques. IEEE access. 2021 Mar 4;9:39707-16.

[10] Bokhare A, Bhagat A, Bhalodia R. Multi-layer Perceptron for Heart Failure Detection Using SMOTE Technique. SN Computer Science. 2023 Jan 25;4(2):182.

[11] Sengupta, Aditya, et al. "Prognostic utility of a novel risk prediction model of 1-year mortality in patients surviving to discharge after surgery for congenital or acquired heart disease." The Journal of Thoracic and Cardiovascular Surgery (2023).

[12] Sakata Y, Shimokawa H. Epidemiology of heart failure in Asia. Circulation Journal. 2013;77(9):2209-17.

[13] Givi M, Sarrafzadegan N, Garakyaraghi M, Yadegarfar G, Sadeghi M, Khosravi A, Azhari AH, Samienasab MR, Shafie D, Saadatnia M, Roohafza H. Persian Registry of cardioVascular diseasE (PROVE): Design and methodology. ARYA atherosclerosis. 2017 Sep;13(5):236.

[14] Kim SI, Noh Y, Kang YJ, Park S, Lee JW, Chin SW. Hybrid data-scaling method for fault classification of compressors. Measurement. 2022 Sep 30;201:111619.

[15] Liu, Mingxuan, et al. "Handling missing values in healthcare data: A systematic review of deep learning-based imputation techniques." Artificial Intelligence in Medicine (2023): 102587.

[16] Josse J, Prost N, Scornet E, Varoquaux G. On the consistency of supervised learning with missing values. arXiv preprint arXiv:1902.06931. 2019 Feb 19.

[17] Le Morvan M, Prost N, Josse J, Scornet E, Varoquaux G. Linear predictor on linearly-generated data with missing values: non consistency and solutions. InInternational Conference on Artificial Intelligence and Statistics 2020 Jun 3 (pp. 3165-3174). PMLR.

[18] Heymans MW, Twisk JW. Handling missing data in clinical research. Journal of clinical epidemiology. 2022 Nov 1;151:185-8.

[19] Saffari SE, Volovici V, Ong ME, Goldstein BA, Vaughan R, Dammers R, Steyerberg EW, Liu N. Proper use of multiple imputation and dealing with missing covariate data. World neurosurgery. 2022 May 1;161:284-90.

[20] Devlin J, Chang MW, Lee K, Toutanova K. Bert: Pre-training of deep bidirectional transformers for language understanding. arXiv preprint arXiv:1810.04805. 2018 Oct 11.

[21] Theerthagiri P. Predictive analysis of cardiovascular disease using gradient boosting based learning and recursive feature elimination technique. Intelligent Systems with Applications. 2022 Nov 1;16:200121.

[22] Liu Z, Song J. Comparison of Tree-based Feature Selection Algorithms on Biological Omics Dataset. InProceedings of the 5th International Conference on Advances in Artificial Intelligence 2021 Nov 20 (pp. 165-169).

[23] Rostami M, Berahmand K, Forouzandeh S. A novel method of constrained feature selection by the measurement of pairwise constraints uncertainty. Journal of Big Data. 2020 Dec;7(1):1-21.

[24] Alghushairy O, Alsini R, Soule T, Ma X. A review of local outlier factor algorithms for outlier detection in big data streams. Big Data and Cognitive Computing. 2020 Dec 29;5(1):1.

[25] Chawla NV, Bowyer KW, Hall LO, Kegelmeyer WP. SMOTE: synthetic minority over-sampling technique. Journal of artificial intelligence research. 2002 Jun 1; 16:321-57.

[26] Batista GE, Bazzan AL, Monard MC. Balancing training data for automated annotation of keywords: a case study. Wob. 2003 Dec 3; 3:10-8.

[27] Kumar, Vinod, et al. "Addressing binary classification over class imbalanced clinical datasets using computationally intelligent techniques." Healthcare. Vol. 10. No. 7. MDPI, 2022.

[28] Chicco D, Oneto L. An enhanced Random Forests approach to predict heart failure from small imbalanced gene expression data. IEEE/ACM Transactions on Computational Biology and Bioinformatics. 2020 Dec 1;18(6):2759-65.

[29] Chen T, Guestrin C. Xgboost: A scalable tree boosting system. InProceedings of the 22nd acm sigkdd international conference on knowledge discovery and data mining 2016 Aug 13 (pp. 785-794).

[30] Yewale D, Vijayaragavan SP, Munot M. An optimized XGBoost based classification model for effective analysis of heart disease prediction. InAIP Conference Proceedings 2023 Jun 23 (Vol. 2768, No. 1). AIP Publishing.

[31] Liu XY, Wu J, Zhou ZH. Exploratory undersampling for class-imbalance learning. IEEE Transactions on Systems, Man, and Cybernetics, Part B (Cybernetics). 2008 Dec 16;39(2):539-50.

[32] Miah, Yunus, et al. "Performance comparison of machine learning techniques in identifying dementia from open access clinical datasets." Advances on Smart and Soft Computing: Proceedings of ICACIn 2020. Springer Singapore, 2021.

[33] Shilaskar S, Ghatol A, Chatur P. Medical decision support system for extremely imbalanced datasets. Information Sciences. 2017 Apr 1;384:205-19.

[34] Brownlee J. Roc curves and precision-recall curves for imbalanced classification. Machine learning mastery. 2020 Jan.